# Towards Data-Driven Automatic Video Editing

Sergey Podlesnyy[1]

[1] Cinema and Photo Research Institute, 47 Leningradsky Ave., Moscow, 126167, Russia
s.podlesnyy@nikfi.ru

**Abstract.** Automatic video editing involving at least the steps of selecting the most valuable footage from points of view of visual quality and the importance of action filmed; and cutting the footage into a brief and coherent visual story that would be interesting to watch is implemented in a purely data-driven manner. Visual semantic and aesthetic features are extracted by the ImageNet-trained convolutional neural network, and the editing controller is trained by an imitation learning algorithm. As a result, at test time the controller shows the signs of observing basic cinematography editing rules learned from the corpus of motion pictures masterpieces.

## 1 Introduction

Portable devices equipped with video cameras, like smartphones and action cameras, show rapid progress in imaging technology. 4K sensor resolution and highly efficient video codecs like H265 provide firm ground for social video capturing and consumption. However, despite this impressive progress, users seem to prefer photos as the main medium to capture their daily events and to consume video clips produced by professionals or skilled bloggers. We argue that the main reason is the time needed to watch the video is orders of magnitude longer than photo browsing. Social video footage captured by ordinary gadget users are too long, not even speaking of quality. They should be edited before presenting them even to closest friends or relatives.

Video editing should involve at least the steps of selecting the most valuable footage from points of view of visual quality and the importance of action filmed. This is a time-consuming process on its own, but the next step is cutting the footage into a brief and coherent visual story that would be interesting to watch. This process in cinematographic and thus requires many artistic and technical skills, which makes it almost impossible for a broad range of users to success.

Since recently, the huge success of deep learning was shown in the visual data processing. We aim to apply the proven techniques of machine learning, convolutional neural networks and reinforcement learning to create automatic tools for social video editing that unifies compression/summarizing and cinematographic approaches for obtaining concise and aesthetically pleasant video clips of memorable moments for a broad audience of unprofessional video gadget users.

**Contributions**: we propose a system capable to learn the editing style from the samples extracted from the content created by professional editors including the motion pictures masterpieces, and to apply this data-driven style to cut unprofessional videos with the ability to mimic the individual style of selected reference samples.

## 2 Related Work

### 2.1 Video Summarizing

A broad literature exists for a video summarizing i.e. automatic rendering of video summary. Although this problem is less directly related to the topic of this paper, we refer to the method described in [1]. They cluster video shots using a measure of visual similarity such as color histograms or transform coefficients. Consecutive frames belonging to the same cluster a considered as video shot. The method of visual clustering is based on low-level graphical measurements and does not contain semantic information about frame content and geometry. In our work [2] we have shown that feature vectors obtained from frame image by a convolutional neural network trained to recognize wide nomenclature if classes, such as ImageNet contest [3] comprise semantic information suitable for visual examples-based information retrieval and for segmenting videos into distinct shots. Here we will show that the same feature vector can be used for automatic video editing.

### 2.2 Automated Editing of Video Footage from Multiple Cameras to Create a Coherent Narrative of an Event

Arev et al [4] describe a system capable of automatic editing of video footage obtained from multiple cameras. Montage is performed by optimizing a path in the trellis graph constructed of frames of multiple sources, ordered in time. The system efficiently produces high-quality narratives by means of constructing the cost functions for nodes and edges of the trellis graph to closely correspond basic rules of cinematography. For example, they estimate 3D camera position and rotation for every source of video footage and further estimate the most important action location in a 3D scene as a joint focus of attention of multiple cameras. By estimating the distance between the camera and joint attention focus the system is capable of evaluating the scale of each shot. Cost functions for the graph edges penalize transitions between the shots more than two sizes apart. Lastly, the system promotes cuts-on-action transitions through actions estimation as local maxima of joint attention focus acceleration. The proposed method relies heavily on the availability of multiple video sources taken from different angles making it possible to derive a joint attention point in 3D space. Cinematography rules are hard-coded into the system.

To prevent including technically defective footage into the resulting film we propose to use the results shown in [5]. They used a crowd-sourced collection of rated photos to train a convolutional neural network to directly regress aesthetical score of an image. It appears that their system penalizes basic technical defects like blurred image, skyline slope, faces occlusions, etc.

### 2.3 Learning editing styles from existing films

As shown in [6], it is possible to model the editing process using a Hidden Markov Model (HMM). They hand-annotated existing film scenes to serve as training data. Concretely, they annotated shot sizes and action types as 'main character entering scene', 'both characters leave scene' etc. But manual annotation is tedious, and this approach still requires access to the underlying scene actions when applying the learned style to new animated scenes.

As can be seen from this review, existing methods of automatic film editing rely on broad meta-data either extracted from multiple footage sources in form of joint attention focus or parsing dialog scripts and even hand-annotating them. In particular, many methods described in the literature require the information about the shot size that is also a cornerstone for traditional cinematography editing rules.

**The rest of this paper is organized as follows.** In Section 3 we share practical details of the early prototype of the system for automatic video editing we are building. Section 4 describes the structure and methods of training of distinct system components, including the shot size classifier and visual quality assessor and reinforcement learning module for video editing control. Section 5 is devoted to the evaluation of results.

## 3 System Overview

Figure 1 shows the video footage features extraction pipeline. Frames are sampled from the video stream of possibly several video files. After simple preprocessing (downscaling to 227x227 pixels, mean color value subtraction) frame images are input into a GPU where a combination of three convolutional neural networks reside. A GoogLeNet [7] network is used to extract a semantic feature vector of length 1024. A network trained to regress aesthetical score on AVA dataset [5] produces a vector of length 2. A network trained to classify an image into three classes of shot sizes (close-up, medium shot, long shot) produces a vector of length 3.

The next step of the pipeline is segmenting video footage into coherent shots as described in [2]: to determine shot boundaries we analyze vector distance between semantic feature vectors of neighboring frames. If vector distance is large we place shot boundary there. For every shot, we calculate the attributes as the mean value of semantic feature vectors and median values of shot size vector and aesthetic score.

We perform the same pipeline for reference cinematography samples as well as for user-generated video content. For reference samples, we obtained DVD copies of 68 out of 100 best movies by The American Society of Cinematographers [8] and processed them excluding 2 minutes of content from the beginning and the end, thus eliminating captions, studio logos, etc. irrelevant material.

The process of automatic film editing works as follows. Features Preparation module reads data from the Shot Attributes database and feeds it into the learned model for automatic video editing. The Editing module produces a storyboard comprising complete shots or their segments from raw footage. Based on that, the Composing module composes the output video clip cutting from raw footage. We are using *ffmpeg* for composing.

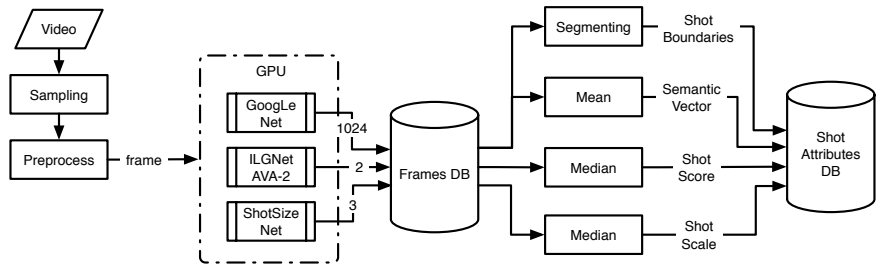

**Fig. 1.** Video footage features extraction pipeline.

## 4 Components and Training

### 4.1 Features Extraction Components

For semantic features extraction, we use GoogLeNet [7] trained by BVLC on the ImageNet dataset. We omit the final classification layer and use the output of layer 'pool5/7x7_s1' as a semantic feature vector of length 1024. After we have extracted feature vectors from over 1,670,000 frames of masterpieces mentioned in Section 3 (a), we performed an incremental PCA to reduce the dimensionality of feature vectors to 64. Residual error for the last batch on incremental PCA was 0.0029.

To automatically detect the shot size, we trained a classifier to distinguish images between 3 classes: close-up, medium shot, long shot. We created a dataset using detailed cinematography scripts of 72 episodes of various Russian TV series. The dataset contained 566,000 frames of nearly even distribution of frames belong to three classes. We trained the GoogLeNet-based network structure with softmax loss after 3-outputs fully connected the final layer. Top-1 testing accuracy was 0.938. We used a trained network published in [5] for the aesthetic scoring of the shot.

### 4.2 Automatic Editing Model Training

We use motion pictures masterpieces as reference samples of good editing. We model video editing as a process of making control decisions whether to include a shot into a final movie or skip it. Further, we model the editing rhythm by learning to make fine-grained decisions of the duration of a shot that was selected to be included in the final movie.

Concretely, we model the video editing process as a sequence learning problem [9] with the Hamming loss function. We selected the following labels for shots sequence labeling (see Table 1).

**Table 1.** Action labels for sequence learning.

| Label | Action |
|-------|--------|
| 1 | Include shot, duration < 1 sec |
| 2 | Include shot, duration 1…3 sec |
| 3 | Include shot, duration 3…9 sec |
| 4 | Include shot, duration > 9 sec |
| 5 | Skip shot |

We prepared the training data as follows. As described in 4.1 we collect a sequence of shots into a clip having a duration of around 2 minutes. Each shot in the clip sequence is labeled according to its duration with labels 1-4 (Table 1). This gives us a reference 'expert' movie cut. To give our model a concept of bad montage, we produce around 40 augmented clips from each reference ones. We do so by randomly inserting shots taken from other movies in masterpieces collection, thus breaking the author's idea of an edit. We assign such a shot a label 5. Additionally, we assign label 5 to all shots having an aesthetical score below some threshold (e.g. 0.1). As a result, we got the training set having 108,491 sample clips, and shuffled the samples. We used *vowpal wabbit* [10] for training in sequence learning task using DAGGER, an iterative algorithm that trains a deterministic policy in imitation learning setting where expert demonstrations of good behavior are used to learn a controller. We construct the enhanced state vector $s$ as follows: we use action labels $a$ in historical span 6, and add neighboring semantic feature vectors from -6th shot to +3rd shot from the current one. Held out loss after 32 epochs of training was 4.06 while the average length of a sequence was 21.

## 5 Evaluation

### 5.1 Qualitative Evaluation

In Fig. 2(a) the test footage is shown in the storyboard format. It contains an unmodified fragment of 'Cool Hand Luke', a 1967 film by Stuart Rosenberg. Fig. 2(b) shows the result of automatic editing. We may observe that the algorithm has shortened the footage and deleted some shots that produced abrupt transitions. For example, the transition between medium I shot and the truck (very long distance shot) is quite abrupt, and since our model learned average rules of editing it removed the truck shot.

Fig. 2(c) shows test footage constructed from the fragment of the movie not seen by the model during training. It is 'Gagarin the First in Space', 2013, editor Pavel Parhomenko. The fragment was augmented by random inserting shots taken from random places of the same movie. In Fig. 2(d) we see the result of auto-editing. It correctly removed the shot with a close-up on the radio but left a few other foreign shots. However overall cut looks smooth and shows a nice gradual change of tone throughout the length of the clip.

In Fig. 2(e) we used a fragment from 'Das Boot', 1981 by Wolfgang Petersen and inserted a few shots from 'Fanny and Alexander', 1982 by Ingmar Bergman, having the tone very much alike the close-up faces of 'Das Boot' but breaking the rule of 180° of action: in 'Das Boot' an officer speaks to a crew in front of him, so the correct cut would be to montage shots with facing directions. Our model correctly removed the wrong shots as shown in Fig. 2(f).

At last, Fig. 2(g) shows the typical unprofessional footage taken by a GoPro camera attached to a bike. Fig. 2(h) demonstrates the result of automatic editing which has resulted in a brief clip showing some sights of dynamism. It contains the main character and beautiful surroundings. A professional editor would probably include shots with the sun shining and a bike wheel close-up. We should leave these artistic enhancements for future work.

## 5.2 Quantitative Evaluation

To estimate how well the proposed method learns basic video editing rules from unlabeled reference samples of motion pictures masterpieces, we have manually counted the number of transitions between the shot sizes in reference footage, in raw unprofessional footage, and in the automatically edited clips. The following shot sizes are distinguished: Detail - Close up - Medium 1 - Medium 2 - Long shot - Very long shot. It was advised, for example by famous Russian cinematographer Kuleshov in the early XX century, that transitions between the shots should occur at two size steps, e.g. between the medium 1 shot and the long shot. Transitions 'Detail - Close up' and 'Long shot - Very long shot' are also allowed.

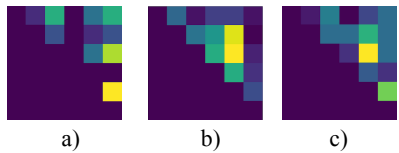

a)          b)          c)

**Fig. 3.** Histograms of the distribution of transitions between the shots. (a) Unprofessional video; (b) Motion Picture Masterpieces; (c) Automatically edited by our algorithm.

In order to evaluate whether the trained DAGGER model has learned the very basic principles of video editing from the data features extracted from motion pictures masterpieces we have calculated the number of transitions between shots of different sizes in three corpuses of footage: clips sampled from masterpieces dataset, clips sampled from unprofessional video footage and clips automatically edited by the DAGGER model. Fig. 3 shows histograms of the distribution of transitions between the shots. It is easy to see that unprofessional video has somewhat even distribution of transitions while motions pictures masterpieces and DAGGER-edited video clips demonstrate a distribution with distinct elevation in the part corresponding to '2 steps difference' between the shot sizes. The standard deviation between masterpieces histogram and unprofessional video histogram is 0.004 while STDEV between masterpieces histogram and clips edited by our algorithm is 0.001 which is 4 times improvement.

# 6 Conclusion and Future Work

These early results make us optimistic to pursue further development of an automatic video editing system capable to mimic the user-selected montage style. We plan to dramatically improve the quality of the resulting video clip by introducing new features into the state vector to give the model clues of the values the users expect from a video. For example, a measure of importance inspired by [1] but constructed from deep learning-based semantic features should promote unique fragments that are most important to preserve. Another line of research here is utilizing the advances in deep reinforcement learning with end-to-end training given visual input data.

## References

1. Uchihachi, S., Foote, J.T., Wilcox, L. US Patent 6,535,639 B1, 2003.
2. Podlesnaya, A., Podlesnyy, S. Deep learning based semantic video indexing and retrieval. In: Proceedings of SAI Intelligent Systems Conference, pp. 359-372.
3. Russakovsky, O., Deng, J., Su, H., Krause, J., Satheesh, S., Ma, S., Huang, Z., Karpathy, A., Khosla, A., Bernstein, M., Berg, A. C., and Fei-Fei, L. ImageNet large scale visual recognition challenge. CoRR, arXiv:1409.0575, 2014.
4. I. Arev, H. S. Park, Y. Sheikh, J. Hodgins, and A. Shamir. Automatic Editing of Footage from Multiple Social Cameras. In: ACM Trans. Graph., 33(4):1–11, July 2014.
5. Xin Jin, Jingying Chi, Siwei Peng, Yulu Tian, Chaochen Ye and Xiaodong Li. Deep Image Aesthetics Classification using Inception Modules and Fine-tuning Connected Layer. The 8th International Conference on Wireless Communications and Signal Processing (WCSP), Yangzhou, China, 13-15 October, 2016.
6. Merabti, B., Christie, M., Bouatouch, K. A Virtual Director Using Hidden Markov Models. In: Computer Graphics Forum, Wiley, 2015, 10.1111/cgf.12775. Hal-01244643.
7. Szegedy, C., Liu, W., Jia, Y., Sermanet, P., Reed, S., Anguelov, D., Erhan, D., Vanhoucke, V., and A. Rabinovich. Going deeper with convolutions. CoRR, arXiv:1409.4842, 2014.
8. ASC Unveils List of 100 Milestone Films in Cinematography of the 20th Century. https://theasc.com/news/asc-unveils-list-of-100-milestone-films-in-cinematography-of-the-20th-century , 2019/01/08.
9. Ross, S.,, Gordon, G.J., Bagnell, J. A. A Reduction of Imitation Learning and Structured Prediction to No-Regret Online Learning. In: Journal of Machine Learning Research - Proceedings Track. 15.
10. Langford, J., Li, L.,Strehl, A. Vowpal wabbit online learning project, 2007. http://hunch.net/?p=309.

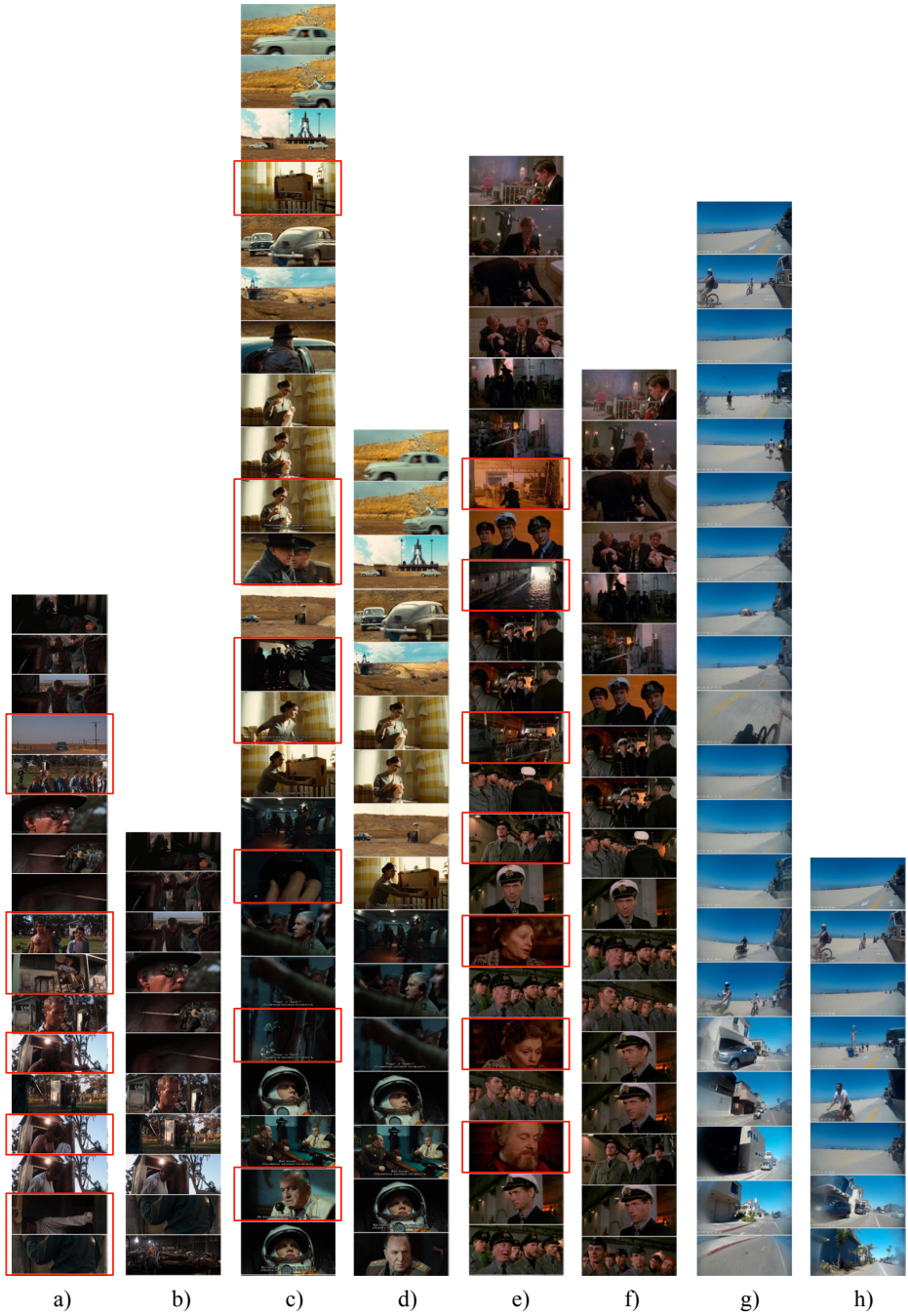

**Fig. 2.** Qualitative results: a, c, e, g – raw materials, b, d, f, h – edited clips. Shots automatically removed from raw materials are marked with a red frame.